\pdfoutput=1

\documentclass[11pt]{article}

\usepackage{acl}
\usepackage{array}
\usepackage{graphicx}
\usepackage{times}
\usepackage{latexsym}
\usepackage{makecell}
\usepackage{subcaption}
\usepackage{amsmath}
\usepackage[T1]{fontenc}

\usepackage[utf8]{inputenc}

\usepackage{microtype}

\usepackage{inconsolata}
\usepackage[skip=0.5\baselineskip]{parskip}

\usepackage{graphicx}
\usepackage{enumitem}
\usepackage[most]{tcolorbox} 
\usepackage{amsmath}
\usepackage{amssymb}
\usepackage{textcomp}
\usepackage{booktabs}
\usepackage[table]{xcolor}
\usepackage{multirow}
\usepackage{graphicx}
\usepackage{caption}
\usepackage{lscape}
\usepackage{makecell}
\definecolor{Block}{RGB}{240,248,255}
\definecolor{rowblue}{HTML}{DAE9F8}

\newcolumntype{P}[1]{>{\centering\arraybackslash}p{#1}}

\DeclareUnicodeCharacter{22A5}{\ensuremath{\bot}} 
\DeclareUnicodeCharacter{00AC}{\ensuremath{\neg}} 
\DeclareUnicodeCharacter{2227}{\ensuremath{\land}} 
\DeclareUnicodeCharacter{2228}{\ensuremath{\lor}} 
\DeclareUnicodeCharacter{2192}{\ensuremath{\rightarrow}} 
\DeclareUnicodeCharacter{2203}{\ensuremath{\exists}} 

\newtcolorbox{bluebox}[1][]{
  enhanced,
  colframe=blue!40!gray,
  colback=white,
  coltitle=white,
  colbacktitle=blue!40!gray,
  width=\linewidth,
  arc=2mm,
  auto outer arc,
  boxrule=0.5pt,
  left=10pt,
  right=10pt,
  drop shadow={black!50!white},
  top=10pt,
  bottom=10pt,
  title={#1}, 
  fonttitle=\bfseries,
  title code={\node[rounded corners, fill=blue!75!black, draw=none, text=white] at (frame.title) {\textbf{#1}};}, 
  attach boxed title to top center={yshift=-2mm},
  boxed title style={sharp corners, size=small}
}

\title{Dissecting Logical Reasoning in LLMs: A Fine-Grained Evaluation and Supervision Study}

\author{%
  Yujun Zhou$^{1,*}$,
  Jiayi Ye$^{2,*}$,
  Zipeng Ling$^{3,*}$,
  Yufei Han$^4$,
  Yue Huang$^1$,
  Haomin Zhuang$^1$\\
  \textbf{
  Zhenwen Liang$^1$
  Kehan Guo$^1$,
  Taicheng Guo$^1$,
  Xiangqi Wang$^1$,
  Xiangliang Zhang$^1$} \\
  $^1$University of Notre Dame $^2$MBZUAI $^3$University of Pennsylvania $^4$INRIA\\ $^*$Equal Contribution\\
  \texttt{\{yzhou25,xzhang33\}@nd.edu}
  }

\begin{document}
\maketitle
\begin{abstract}
Logical reasoning is a core capability for large language models (LLMs), yet existing benchmarks that rely solely on final-answer accuracy fail to capture the quality of the reasoning process. To address this, we introduce FineLogic, a fine-grained evaluation framework that assesses logical reasoning across three dimensions: overall accuracy, stepwise soundness, and representation-level probing. Leveraging this framework, we conduct a comprehensive study on how different supervision formats in fine-tuning shape reasoning abilities. We fine-tune LLMs on four supervision styles—one in natural language and three symbolic variants—and find a key trade-off: natural language supervision excels at generalization to out-of-distribution and long-chain problems, whereas symbolic supervision is superior at instilling structurally sound, atomic reasoning steps. Furthermore, our probing analysis indicates that fine-tuning primarily refines the model's step-by-step generation process, rather than improving its ability to converge on an answer early. Together, our framework and analysis provide a more rigorous lens for evaluating and improving logical reasoning in LLMs. The code is available at \url{https://github.com/YujunZhou/FineLogic}.
\end{abstract}

\section{Introduction}

\begin{figure*}[t]
    \centering
    \includegraphics[width=0.92\linewidth]{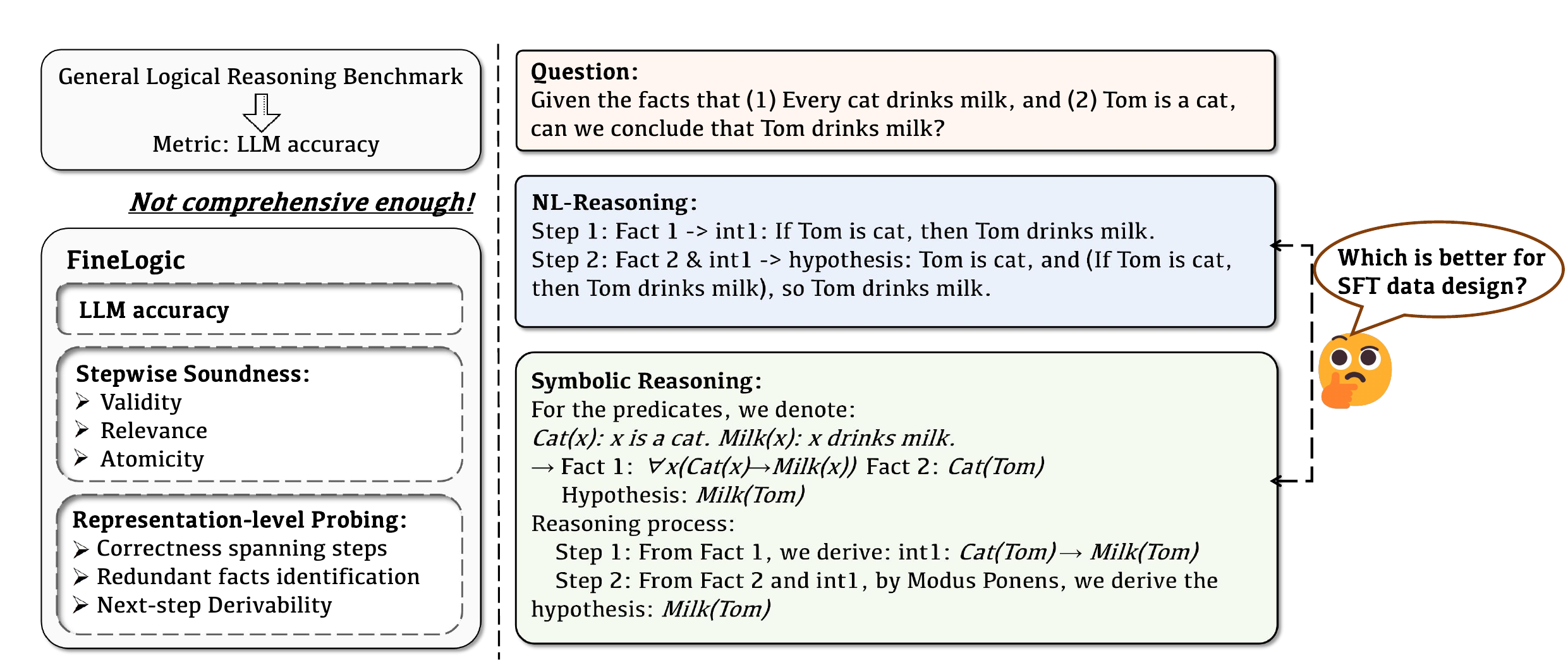} 
    \caption{(Left) LLM logical reasoning evaluation:   the general benchmark v.s. our fine-grained benchmark \textbf{FineLogic}. (Right)  processing a logical reasoning task using natural language v.s. using symbolic methods.}
    \label{fig:figure_page_2}
\end{figure*}

Large language models (LLMs) are rapidly emerging as transformative tools across a wide array of applications \cite{achiam2023gpt, guo2024large, thirunavukarasu2023large,nam2024using,huang2024social,huang2025trustworthiness,zhou2024defending}. Among these, reasoning serves as a core capability underpinning tasks such as problem-solving \cite{lu2023mathvista}, scientific question answering \cite{guo2024can,zhou2024labsafety}, and code analysis \cite{nam2024using}. Consequently, a growing body of research has sought to evaluate and enhance the reasoning abilities of LLMs from multiple perspectives \cite{wei2022chain,guo2025deepseek,guo2024can,liang2024scemqa,wang2025adareasoner,zhou2025evolvinglanguagemodelslabels,zhao2025one}.
Within this broader landscape, logical reasoning stands out as a particularly challenging and intellectually demanding domain \cite{saparov2022language}. It requires a synthesis of natural language understanding, formal logical interpretation, and multi-step inferential processing \cite{patel2024multi, saparov2023testing, morishita2024enhancing}. 

Despite growing interest in the logical reasoning capabilities of LLMs, most existing benchmarks focus narrowly on whether a model produces the correct final answer \cite{patel2024multi,parmar2024logicbench,han2022folio}. This binary evaluation, typically assessing only the correctness of a “True” or “False” output, can be misleading, as it fails to determine whether the model arrived at the answer through valid multi-step reasoning \cite{saparov2022language}. Consequently, correct answers may reflect guesswork rather than genuine logical inference. We are thus motivated  to address 
\textbf{RQ1: How to rigorously evaluate LLMs’ step-by-step correctness in logical reasoning tasks, beyond the binary evaluation of the final answer?}

In parallel with benchmarking efforts, numerous methods have been proposed to enhance the multi-step logical reasoning abilities of LLMs. While many
leverage inference-time strategies \cite{wang2025stepwise}, in-context learning \cite{creswell2022selection, xu2024faithful}, or external logical verifiers \cite{pan2023logic} to guide the model toward more rigorous reasoning, 
some recent studies explored supervised fine-tuning (SFT) as a more direct approach to enhancing logical reasoning \cite{morishita2024enhancing, feng2023language}. For example, \citet{morishita2024enhancing} proposes a synthetic logic corpus designed to offer broad and systematic coverage of logical knowledge. However, it remains unclear for this important question, 
\textbf{RQ2: What style of training data, natural language or formal logical symbols, better facilitates the learning of multi-step logical reasoning through SFT?}
Addressing this research question is important for understanding how to most effectively instill logical reasoning capabilities in LLMs.

To address RQ1, we propose \textbf{FineLogic}, a diagnostic framework designed to offer a multi-dimensional evaluation of LLMs' reasoning processes, moving beyond binary final-answer correctness. Rather than creating another leaderboard, FineLogic serves as a tool for LLM practitioners to pinpoint specific weaknesses in reasoning chains, such as flawed logic, redundancy, or non-atomic steps. Our framework evaluates models along three complementary dimensions: (1) \textbf{Overall benchmark accuracy}: This metric captures a model's ability to perform multi-step logical reasoning and its generalizability across problems from diverse domains.
(2) \textbf{Stepwise Soundness}: Inspired by \citet{saparov2022language}, we assess the quality of each intermediate reasoning step using three criteria—\textbf{validity} (whether the step is logically valid), \textbf{relevance} (whether its conclusion is used in later steps), and \textbf{atomicity} (whether it applies a single, minimal inference rule). These metrics aim to evaluate the model’s ability to generate human-interpretable and logically coherent reasoning chains.
(3) \textbf{Representation-level probing} \cite{ye2024physics}: 
By applying probing techniques to LLM hidden representations, this evaluation provides insight into whether the model's understanding of logical structure is merely surface-level or embedded in its internal state.

To address RQ2, we systematically investigate how different supervision formats affect the reasoning capabilities of LLMs. Specifically, we examine both natural language-based training data and logic-symbol-based representations, including several structured variants.
Our analysis shows that \textbf{natural language supervision is particularly effective in conveying core reasoning patterns}, leading to strong performance across a wide range of evaluation benchmarks. Notably, it exhibits impressive \textbf{generalizability even on out-of-distribution test sets} that require long reasoning chains. However, a deeper examination of \textit{stepwise soundness} and \textit{internal representation probing} reveals certain limitations. Models trained with natural language supervision tend to struggle with producing strictly minimal reasoning chains (e.g.,  more likely including \textbf{redundant steps} and applying multiple inference rules in a single step, as shown in \autoref{fig:casestudy_ness}). In contrast, models trained with symbolic reasoning styles are better at filtering out irrelevant information, generating atomic steps aligned with individual deduction rules, and maintaining cleaner, logically grounded reasoning trajectories. 

To summarize, our contributions are as follows:
\begin{itemize}[leftmargin=12pt]
\item We propose FineLogic, a unified diagnostic framework for assessing LLMs' logical reasoning. It moves beyond final-answer accuracy to evaluate the quality and coherence of reasoning steps, offering fine-grained insights that can guide targeted model improvements and serve as a scaffold for structured reward design in reinforcement learning.

\item We conduct a comprehensive study on the effects of supervision format, fine-tuning LLMs on both natural language and symbolic logic data to examine their impact on reasoning across general and complex tasks.

\item Through systematic analysis of models trained with different supervision styles, we identify key trade-offs between generalization and structural reasoning quality. The findings provide concrete insights into the design and selection of effective training data for post-training.
\end{itemize}

\section{Related Works}
\textbf{Logical Reasoning Benchmarks.} Numerous benchmarks have been proposed to evaluate the logical reasoning abilities of LLMs. Many either mix logical and commonsense reasoning \cite{liu2023logiqa,luo2023towards, havrilla2024glore}, making it hard to isolate logical competence, or assess multi-step reasoning using only final-answer accuracy \cite{parmar2024logicbench,han2022folio,tafjord2020proofwriter,mondorf2024liar}. While ProntoQA \cite{saparov2022language,saparov2023testing} pioneered stepwise evaluation, its analysis is confined to step correctness on short problems. FineLogic moves beyond this by introducing a more comprehensive suite of step-level diagnostics—including a more practical relevance metric—to rigorously assess complex, long-chain reasoning. Furthermore, we are the first to adapt representation-level probing from mathematics \cite{ye2024physics} to the logical domain, introducing novel tasks like Correctness Spanning Steps (CSS) to connect behavioral outputs with internal model states. In contrast, other stepwise frameworks like ROSCOE \cite{golovneva2022roscoe} and RECEVAL \cite{prasad2023receval} remain too generic or coarse-grained for the specific demands of formal logic.

\begin{figure*}[t]
    \centering
    \includegraphics[width=\linewidth]{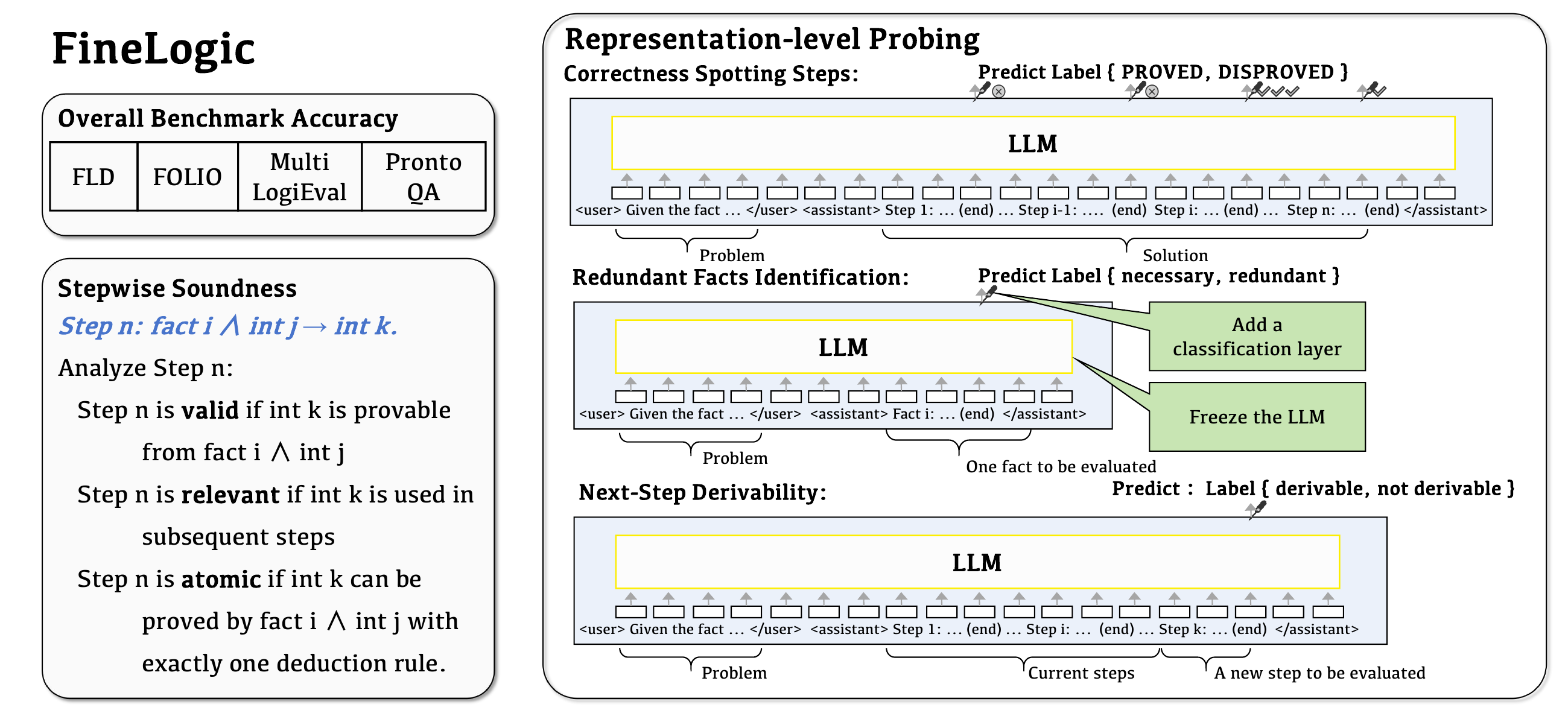} 
    \caption{Overview of FineLogic, where overall benchmark accuracy, stepwise soundness, and representation-level probing are combined for a fine-grained evaluation of LLM's logical reasoning ability.}
    \label{fig:finelogic_section3}
\end{figure*}

\textbf{Logical Reasoning Enhancement.} Several studies have aimed to improve LLMs’ performance on logical reasoning tasks. Some approaches rely on translating inputs into formal logic and using programmable verifiers to solve problems \cite{olausson2023linc,pan2023logic,yang2023harnessing, ryu2024divide}, which bypasses the model’s own reasoning process. Others use in-context learning or inference-time strategies to guide output without fundamentally enhancing reasoning ability \cite{creswell2022selection,wang2025stepwise,xu2024faithful,sun2023determlr,toroghi2024verifiable}. While a few works have explored fine-tuning or reinforcement learning to strengthen logical reasoning \cite{feng2023language,morishita2023learning,morishita2024enhancing,xie2025logic,yang2022generating,xie2024memorization,zheng2025learning}, they have not examined which types of supervision are most effective for teaching LLMs to reason. In this work, we focus specifically on this open question.

\section{FineLogic Evaluation Framework}

As illustrated in Figure~\ref{fig:finelogic_section3}, FineLogic builds on existing benchmarks and evaluates logical reasoning ability from three complementary perspectives:
(1) \textbf{Overall benchmark accuracy}, which measures whether the model can correctly solve multi-step reasoning tasks;
(2) \textbf{Stepwise soundness}, which evaluates whether each reasoning step is valid and interpretable;
(3) \textbf{Representation-level probing}, which assesses whether the model internally captures the problem’s reasoning structure beyond surface-level patterns.

\subsection{Overall Benchmark Accuracy}
Similar to most benchmarks, our overall benchmark accuracy focuses on final-answer correctness. Specifically, we define accuracy (Acc) as:
{\begin{equation}
    \text{Acc} = \frac{1}{N} \sum_{i=1}^{N} \mathbf{1}[\hat{y}_i = y_i]
\end{equation}}

where $N$ is the number of test problems, $y_i$ is the gold label, and $\hat{y}_i$ is the model's prediction for problem $i$. While coarse-grained, it offers a quick and effective way to assess a model’s overall reasoning ability and cross-domain generalization. We evaluate on four challenging multi-step reasoning benchmarks, deliberately selected for their focus on peer-reviewed, multi-hop formal logic. The suite includes widely-used datasets for comparability (\textbf{FOLIO} \cite{han2022folio} and \textbf{ProntoQA} \cite{saparov2022language}) alongside benchmarks designed to probe out-of-distribution generalization by varying complexity and rule composition (\textbf{FLD} \cite{morishita2024enhancing} and \textbf{Multi-LogiEval} \cite{patel2024multi}). This selection offers a comprehensive assessment, in contrast to narrower benchmarks like ProofWriter \cite{tafjord2020proofwriter}, which we exclude due to its limited scope of logical rules. Further details on dataset statistics and sampling are provided in Table~\ref{tab:dataset_samples} and Appendix~\ref{app:datasets}.

\begin{table}[t]
  \centering
  \resizebox{1\linewidth}{!}{
\begin{tabular}{lcc}
\toprule
\textbf{Dataset} & \textbf{Samples} & \textbf{Label Types} \\
\midrule
FLD \cite{morishita2024enhancing}       & 1100 & \{T, F, Unknown\} \\
FOLIO \cite{han2022folio}               & 203  & \{T, F, Unknown\} \\
Multi-Logical \cite{patel2024multi}     & 390  & \{T, F\} \\
Pronto-QA \cite{saparov2022language}    & 500  & \{T, F\} \\
\bottomrule
\end{tabular}
  }\vspace{-0.1in}
  \caption{Sample counts and label types for each dataset.} \vspace{-0.2in}
  \label{tab:dataset_samples}
\end{table}

\subsection{Stepwise Soundness}
Building on \citet{saparov2022language}, we evaluate the soundness of each intermediate reasoning step along three dimensions: \textbf{validity} (whether the step logically follows from its premises), \textbf{relevance} (whether its conclusion is used in later steps), and \textbf{atomicity} (whether it applies a single, minimal inference rule).

To assess these criteria, we extract the premises and conclusion of each step from the model's final answer. Crucially, our evaluator scores only the text following a designated answer tag (e.g., </think>), ensuring that any preceding self-reflection or speculative reasoning is excluded by design. We use GPT-4.1-mini to evaluate \textit{validity} and \textit{atomicity}. Manual verification on 200 annotated steps shows that GPT-4.1-mini achieves over 98\% accuracy on both metrics. For \textit{relevance}, we determine whether the conclusion of step $i$ is referenced in any subsequent step $k > i$.

We then compute the proportion of samples in which \emph{all} steps are valid, relevant, and atomic, providing a sample-level measure of reasoning integrity. Formally, for each solution $s$ with $K_s$ steps, let $v_{s,k}, r_{s,k}, a_{s,k} \in \{0, 1\}$ denote the validity, relevance, and atomicity of step $k$, respectively. We define the sample-level metrics as:
{\begin{align}
        \text{AllValid} &= \frac{1}{N} \sum_{s=1}^{N} \left[ \prod_{k=1}^{K_s} v_{s,k} \right] \quad \\
\text{AllRelevant} &= \frac{1}{N} \sum_{s=1}^{N} \left[ \prod_{k=1}^{K_s} r_{s,k} \right] \quad \\
\text{AllAtomic} &= \frac{1}{N} \sum_{s=1}^{N} \left[ \prod_{k=1}^{K_s} a_{s,k} \right]
\end{align}}
Full prompt templates are provided in Figures~\ref{prompt:StepValidity} and~\ref{prompt:StepAtomicity}.

\subsection{Representation-level Probing}
Inspired by \citet{ye2024physics}, we introduce \textbf{representation-level probing accuracy} to assess whether LLMs internally understand how and when to perform specific reasoning step. Unlike behavioral metrics, this method aligns internal representations with reasoning structure and tracks how reasoning knowledge evolves across steps.

We construct probing datasets from FLD test samples requiring 10–20 reasoning steps, using 450 problems for training and 100 for testing across three tasks, implementation details are provided in Appendix~\ref{app:probing}:

\textbf{Correctness Spanning Steps (CSS)}: Identifies the earliest step after which the model consistently predicts the correct label. The spanning length is the number of remaining steps from that point to the end. Higher accuracy indicates earlier internalization of the correct answer. Formally, for each problem $s$, let $\tau_s$ be the first step from which the probe consistently predicts the correct label and $K_s$ be the total number of steps. The CSS score is then:
\begin{equation}
    \text{CSS} = \frac{1}{N} \sum_{s=1}^{N} (K_s - \tau_s).
\end{equation}

\textbf{Redundant Facts Identification (RFI)}: After presenting all facts and the hypothesis, we append three necessary and three redundant facts. A classifier is trained to distinguish between them, measuring the model’s ability to identify irrelevant information. Higher accuracy reflects better fact discrimination.

\textbf{Next-Step Derivability (NSD)}: At six randomly selected intermediate steps, we append three valid and three invalid candidate steps. Probing predicts which are currently derivable. Higher accuracy indicates stronger awareness of valid next steps.

For both RFI and NSD, performance is measured using balanced accuracy to account for the construction of positive and negative instances. The score is calculated as:
$$
\text{Score} = \frac{1}{2}(\text{TPR} + \text{TNR}),
$$
where TPR is the True Positive Rate and TNR is the True Negative Rate.

Our evaluation builds on two prior lines of work—stepwise reasoning evaluation \cite{saparov2022language} and representation-level probing \cite{ye2024physics}—but introduces key extensions tailored to logical reasoning.



\section{Supervision Format and Style: SFT Data Design}
\label{sec:sft_data_design}

In this section, we examine how different supervision styles for SFT affect the logical reasoning abilities of LLMs. Our training data is based on \textbf{FLD} and \textbf{ProntoQA}, both of which include gold reasoning chains suitable for constructing diverse supervision styles.

For \textbf{FLD}, we generate 500 problems for each reasoning depth from 0 to 15, plus 1500 \texttt{UNKNOWN} samples, totaling 9500 training instances. For \textbf{ProntoQA}, we use 3200 3-hop problems. During evaluation, FLD covers depths 0–19, while ProntoQA uses only the hardest 5-hop samples.

We compare four supervision styles across two categories: natural language-based and symbolic reasoning. Each style reflects a different level of abstraction and clarity in reasoning structure.

\begin{itemize}[leftmargin=12pt]
\item \textbf{NL-Reasoning}: Solutions are written entirely in natural language, with no intermediate symbolization or abstraction.

\item \textbf{Symbolic Reasoning (Structured)}: Problems are formalized by defining variables and predicates, translating facts and hypotheses into logical forms, and reasoning step by step using symbolic logic.

\item \textbf{Symbolic Reasoning (Filtered)}: A simplified variant where only necessary facts are retained, shortening reasoning chains and reducing input complexity.

\item \textbf{Symbolic Reasoning (Direct)}: Facts are directly expressed in symbolic form without defining variables or predicates, which shortens sequences but may introduce ambiguity.
\end{itemize}

A small portion of translations, connective phrases, and intermediate steps are generated using GPT-4.1. Prompt examples are shown in Figure~\ref{fig:different_training_case} (Appendix~\ref{app:training_examples}).
\vspace{-3pt}
\section{Experiments}
\subsection{Experimental Setup}

Our experiments involve four base models for fine-tuning—\textbf{LLaMA-3.1-8B-Instruct}, \textbf{Qwen-2.5-7B-Instruct}, \textbf{Qwen-2.5-Math-7B-Instruct}, and \textbf{Qwen-3-8B}—and two powerful zero-shot baselines, \textbf{GPT-4o} and \textbf{DeepSeek R1}. The four base models are fine-tuned for 3 epochs at a $1 \times 10^{-6}$ learning rate, with their original versions also serving as baselines. Representation-level probing is limited to the LLaMA and Qwen models due to computational constraints, and an explicit step-by-step format is enforced for all stepwise evaluations.

We compare SFT models trained with different supervision styles against these baselines:

\begin{itemize}[left=2pt,itemsep=2pt,parsep=0pt]  
\setlength{\leftmargin}{0pt}  
\setlength{\itemindent}{0pt}  
\item \textbf{Direct Answering}
\item \textbf{Chain-of-Thought (CoT)} \citep{wei2022chain}
\item \textbf{Few-Shot Learning} \cite{brown2020language}
\item \textbf{LOGIPT} \citep{creswell2022selection}
\item \textbf{Selection-Inference} \citep{creswell2022selection}
\item \textbf{SymbCoT} \cite{xu2024faithful} 
\item \textbf{LogicLM} \cite{pan2023logic} 
\end{itemize}
More detailed experimental setups can be found in Appendix~\ref{app:detailed_setups}.

\begin{table*}[t]
\centering
\renewcommand{\arraystretch}{1}
\scalebox{0.75}{%
\begin{tabular}{cc P{2.2cm} P{2.2cm} P{2.2cm} P{2.2cm}}
\toprule
\multirow{2}{*}{\textbf{Model}} & \multirow{2}{*}{\textbf{Setting}} & \multirow{2}{*}{\textbf{FLD}} & \multirow{2}{*}{\textbf{FOLIO}} & \multirow{2}{*}{\makecell[c]{\textbf{Multi-}\\\textbf{LogiEval}}} & \multirow{2}{*}{\makecell[c]{\textbf{ProntoQA}}} \\
& & & & & \\
\midrule
\rowcolor{gray!10}
& Direct     & 53.0   & 72.4 & 71.0   & 98.8 \\
\rowcolor{gray!10}
& CoT        & 54.1 & 69.5 & 76.9 & 98.6 \\
\rowcolor{gray!10}
& Few-shot   & \textbf{58.3} & \textbf{74.4} & 84.4 & 99.0 \\
\rowcolor{gray!10}
& Logic-LM   & 46.9 & 72.1 & 83.3 & \textbf{100} \\
\rowcolor{gray!10}
& SymbCoT    & 47.6 & 71.6 & 72.1 & \textbf{100} \\
\rowcolor{gray!10} \multirow{-6}{*}{\textbf{GPT-4o}}
& Sel-Inf    & 51.9 & 66.5 & \textbf{84.9} & 94.4 \\
\addlinespace
& Direct     & 77.2 & 75.9 & 81.8 & \textbf{100} \\
& CoT        & 77.6 & 78.8 & 79.0 & \textbf{100} \\
& Few-shot   & 77.3 & 81.8 & \textbf{84.6} & 99.4 \\
& Logic-LM   & 69.6 & 77.5 & 81.2 & 96.4 \\
& SymbCoT    & 69.6 & 82.8 & 72.0 & 98.2 \\ \multirow{-6}{*}{\textbf{\shortstack{DeepSeek-R1}}}
& Sel-Inf    & \textbf{83.8} & \textbf{85.2} & 73.1 & 96.0 \\
\addlinespace
\rowcolor{gray!10}
& Direct     & 31.7 & 54.7 & 40.5 & 64.6 \\
\rowcolor{gray!10}
& CoT        & 29.3 & 50.7 & 44.6 & 63.8 \\
\rowcolor{gray!10}
& Few-shot   & 41.0 & 46.5 & 59.4 & 48.9 \\
\rowcolor{gray!10}
& Logic-LM   & 38.3 & 52.5 & 44.4 & 77.6 \\
\rowcolor{gray!10}
& SymbCoT    & 38.1 & 58.8 & 46.3 & 78.8 \\
\rowcolor{gray!10}
& Sel-Inf    & 48.5 & 47.5 & 55.2 & 64.2 \\
\rowcolor{gray!10}
& LogiPT     & 53.3 & \textbf{61.7} & 57.9 & 76.4 \\
\rowcolor{gray!10}
& SFT-NL     & \textbf{67.5} & 57.1 & \textbf{71.3} & 99.6 \\
\rowcolor{gray!10}
& SFT-Symb-Struct & 63.2 & 56.2 & 59.7 & \textbf{99.8} \\
\rowcolor{gray!10}
& SFT-Symb-Filter & 66.7 & 54.7 & 50.8 & 91.0 \\
\rowcolor{gray!10} \multirow{-11}{*}{\textbf{\shortstack{Llama-3.1-8B-Instruct}}}
& SFT-Symb-Direct & 52.8 & 48.3 & 53.9 & 98.8 \\
\addlinespace
& Direct     & 46.6 & 61.1 & 37.0 & 90.6 \\
& CoT        & 50.4 & 65.5 & 54.3 & 90.4 \\
& Few-shot   & 53.2 & 68.5 & 61.3 & 91.1 \\
& Logic-LM   & 46.6 & \textbf{69.1} & 27.1 & 85.8 \\
& SymbCoT    & 22.6 & 57.5 & \textbf{63.9} & 87.0 \\
& Sel-Inf    & 49.0 & 62.6 & 39.7 & 92.6 \\
& LogiPT     & 58.6 & 61.7 & 55.6 & 52.4 \\
& SFT-NL     & \textbf{71.0} & 62.6 & 64.3 & \textbf{97.4} \\
& SFT-Symb-Struct & 54.6 & 50.7 & 57.7 & 83.8 \\
& SFT-Symb-Filter & 54.7 & 55.7 & 61.0 & 96.0 \\ \multirow{-11}{*}{\textbf{\shortstack{Qwen-2.5-7B-Instruct}}}
& SFT-Symb-Direct & 54.8 & 53.2 & 58.7 & 61.4 \\
\bottomrule
\end{tabular}
}
\caption{Overall Benchmark Accuracy on four models with different settings.}
\label{tab:main_results}
\end{table*}


\begin{figure*}[t]
    \centering
    \begin{subfigure}[t]{0.49\linewidth}
        \centering
        \includegraphics[width=\linewidth]{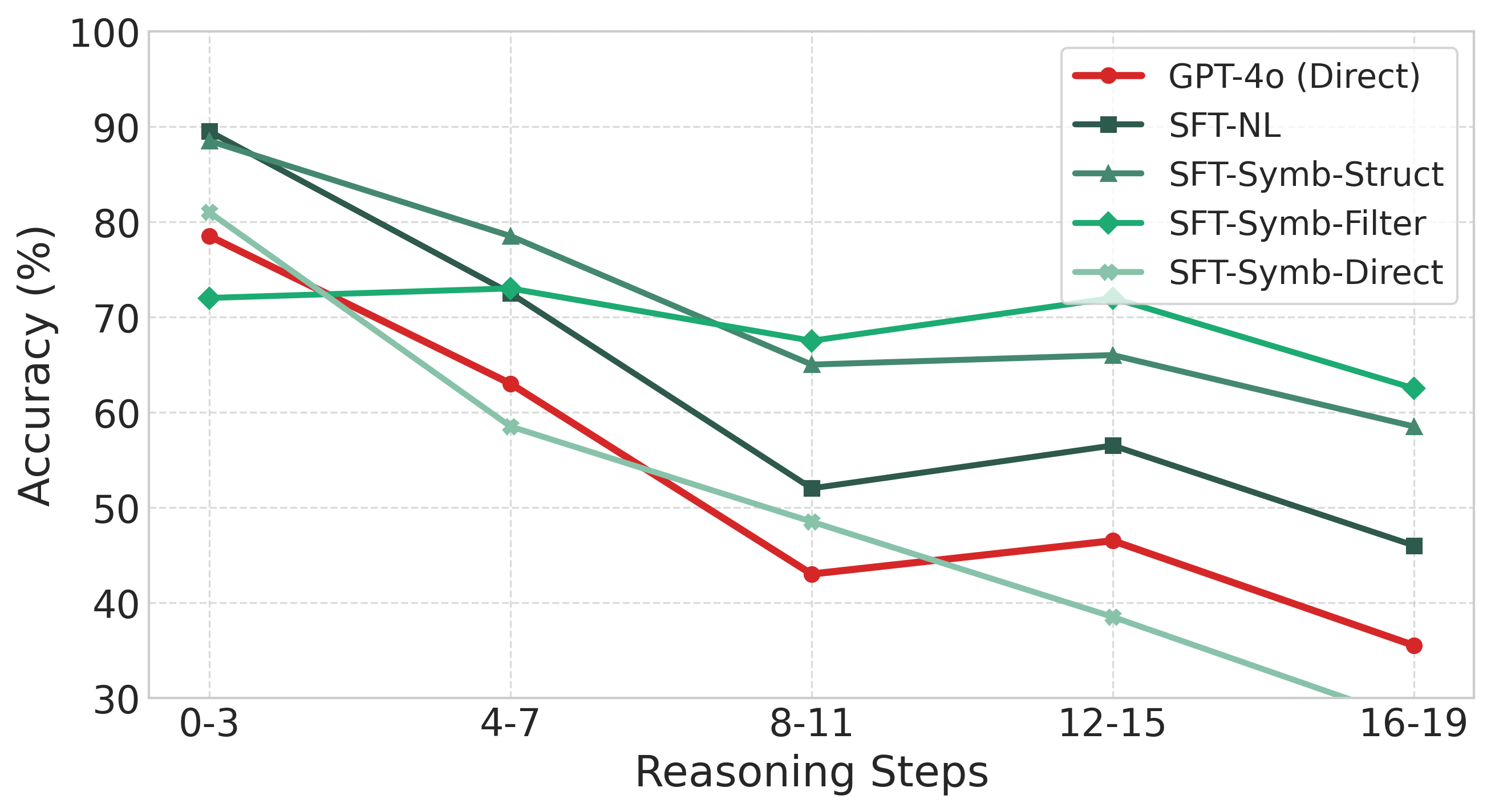}
        \caption{Performance of Llama-3.1-8B-Instruct SFT.}
        \label{fig:llama_steps}
    \end{subfigure}%
    \hspace{0.01\linewidth}
    \begin{subfigure}[t]{0.49\linewidth}
        \centering
        \includegraphics[width=\linewidth]{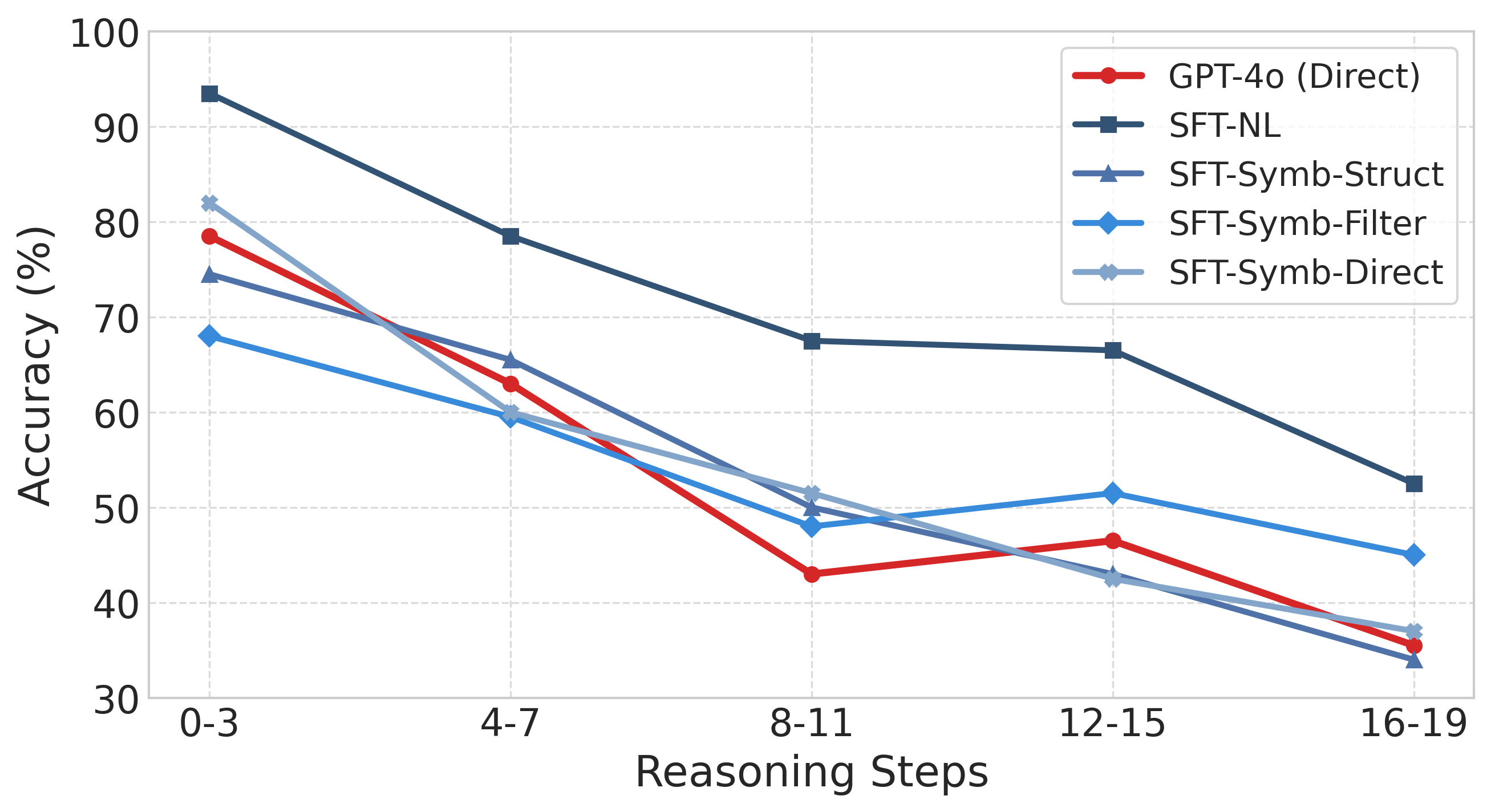}
        \caption{Performance of Qwen-2.5-7B-Instruct SFT.}
        \label{fig:qwen_steps}
    \end{subfigure}
    \caption{Comparison of SFT variants' performance across different reasoning step ranges in FLD dataset. Both charts show accuracy declines with increasing inference steps, with GPT-4o (Direct) included as a reference. In (a), Llama with SFT-Symb-Filter maintains strong performance even in the 16-19 step range (out-of-distribution), while in (b), Qwen with SFT-NL shows remarkable early-stage reasoning capabilities. }
    \label{fig:step_performance}
\end{figure*}

\subsection{Results}
We conducted experiments for analyzing the performance of four models combined with various prompting and fine-tuning settings under the \textbf{FineLogic Evaluation Framework}. Due to space constraints, the results for Qwen-2.5-Math-7B-Instruct and Qwen-3-8B are presented in Appendix~\ref{app:experiment_details}.

\subsubsection{Results on Overall Benchmark Accuracy }
As shown in \autoref{tab:dataset_samples}, we report the \textbf{overall benchmark accuracy} across four datasets, as well as the \textbf{step-wise accuracy} on the FLD benchmark, stratified by reasoning depth (Figure \ref{fig:step_performance}).
Our analysis yields several key observations:

\textbf{CoT and few-shot prompting generally improve performance, but baseline methods do not consistently yield gains.}
Across the four evaluation datasets, both CoT and few-shot prompting lead to broadly positive improvements, indicating their general effectiveness in enhancing LLM performance on logical reasoning tasks. Notably, \textbf{few-shot prompting consistently outperforms CoT}, suggesting that for complex logical tasks, showing the model \emph{how} to think (via exemplars) is more beneficial than simply encouraging it to reason step by step. This may be because logical questions naturally elicit multi-step reasoning under direct prompting, limiting the marginal benefit of CoT. In contrast, few-shot demonstrations provide clearer procedural scaffolding, which appears more effective in guiding the model's reasoning process.

In contrast, baseline prompting methods such as \textit{Logic-LM}, \textit{SymbCoT}, and \textit{Sel-Inf} show inconsistent performance and sometimes underperform even direct prompting. For example, \textit{Logic-LM} performs well on simpler problems but degrades on complex ones, with Qwen’s Multi-LogiEval accuracy dropping to 27.1\%. \textit{SymbCoT} sometimes improves over \textit{Logic-LM} (e.g., 63.8\% on Multi-LogiEval with Qwen) but also shows large drops elsewhere (e.g., 22.6\% on FLD, versus 44.6\% with direct prompting).

\textbf{Supervised fine-tuning outperforms inference-time methods, but its effectiveness heavily depends on the supervision style.}
Compared to inference-time prompting strategies, SFT yields significantly greater improvements in logical reasoning performance. Among all training styles, \textbf{natural language-based supervision (SFT-NL) produces the most substantial and consistent gains across datasets and models.}

Notably, even though SFT was conducted using only problems from FLD and ProntoQA with reasoning depths \emph{less than those in the test set}, the resulting models show robust improvements. For example, under the SFT-NL setting, Llama's accuracy on FLD increased from 31.7\% (direct prompting) to 67.5\% and Qwen improved from 46.6\% to 71.0\%, approaching the best-performing baseline DeepSeek R1. On ProntoQA, most SFT variants achieve over 90\% accuracy. Furthermore, even on out-of-distribution datasets such as FOLIO and Multi-LogiEval, some SFT settings deliver strong generalization. For instance, on Multi-LogiEval, Llama with SFT-NL improved to 71.3\%, matching the performance of GPT-4o.

While SFT-NL demonstrates the best overall and most transferable performance, other styles of supervision yield much smaller gains. This may be since LLMs are primarily pretrained on natural language data, making symbolic reasoning—especially when it requires both translation and inference over logic forms—significantly more challenging. Among the symbolic settings, \textbf{SFT-Symb-Filter} consistently outperforms other variants. By removing redundant reasoning steps from the symbolic training data, this setting simplifies training and enhances performance. In contrast, \textbf{SFT-Symb-Direct}, which skips variable and predicate definitions entirely, performs poorly, likely due to the introduction of ambiguity and the lack of explicit logical structure.

\textbf{Accuracy declines with reasoning depth, but SFT enables small models to match GPT-4o even on the most challenging out-of-distribution samples.}
As shown in Figure \ref{fig:step_performance}, model accuracy decreases as the required number of reasoning steps increases. Nonetheless, our results show that SFT substantially improves model robustness, even on long-chain, out-of-distribution examples. On in-distribution FLD test problems (0–15 steps), SFT models trained under most styles outperform GPT-4o. For instance, across reasoning depths up to 15, both Llama and Qwen with SFT-NL surpass GPT-4o's performance.

On more difficult out-of-distribution questions requiring 16–19 steps of reasoning—where no training samples are available—performance drops by approximately 10\% relative to the 12–15 step range. However, even under these conditions, SFT models maintain accuracy comparable to GPT-4o. Combined with strong generalization to unseen datasets such as FOLIO and Multi-LogiEval, these results suggest that \textbf{SFT induces genuine logical reasoning ability in LLMs}. At the same time, the sharp performance decline on longer reasoning chains implies that some portion of success on shorter problems may still stem from shallow pattern matching or memorization, rather than robust inference. Detailed results can be found in \ref{app:experiment_details}.

\subsubsection{Results on Stepwise Soundness}
\begin{table}[t]
\centering
\resizebox{\linewidth}{!}{%
\renewcommand{\arraystretch}{1.15}
\setlength{\tabcolsep}{6pt}
\begin{tabular}{l|lrrr}
\toprule
\textbf{Model} & \textbf{Setting} & \textbf{All Valid} & \textbf{All Relevant} & \textbf{All Atomic}\\
\midrule
GPT-4o          & \cellcolor{rowblue}Few-shot        & \cellcolor{rowblue}7.6  & \cellcolor{rowblue}\textbf{56.2} & \cellcolor{rowblue}4.4 \\\midrule
Deepseek-R1     & Few-shot                           & 13.1 & 33.8 & 5.7 \\\midrule

\multirow{6}{*}{\centering \shortstack{Llama-3.1-\\8B-Instruct}}
                & \cellcolor{rowblue}Few-shot        & \cellcolor{rowblue}4.5  & \cellcolor{rowblue}17.4 & \cellcolor{rowblue}1.6 \\
                & LogiPT                             & 5.2  & \textbf{28.5} & 4.9 \\
                & \cellcolor{rowblue}SFT-NL          & \cellcolor{rowblue}\textbf{40.9} & \cellcolor{rowblue}8.5  & \cellcolor{rowblue}13.0 \\
                & SFT-Symb-Struct                    & 35.0 & 15.4 & 24.7 \\
                & \cellcolor{rowblue}SFT-Symb-Filter & \cellcolor{rowblue}21.8 & \cellcolor{rowblue}16.9 & \cellcolor{rowblue}12.4 \\
                & SFT-Symb-Direct                    & 33.7 & 10.2 & \textbf{25.1} \\\midrule

\multirow{6}{*}{\centering \shortstack{Qwen-2.5-\\7B-Instruct}}
                & \cellcolor{rowblue}Few-shot        & \cellcolor{rowblue}10.1 & \cellcolor{rowblue}35.1 & \cellcolor{rowblue}2.6 \\
                & LogiPT                             & 6.4  & \textbf{39.8} & 5.3 \\
                & \cellcolor{rowblue}SFT-NL          & \cellcolor{rowblue}27.6 & \cellcolor{rowblue}5.4  & \cellcolor{rowblue}8.5 \\
                & SFT-Symb-Struct                    & \textbf{35.3} & 9.1  & \textbf{19.8} \\
                & \cellcolor{rowblue}SFT-Symb-Filter & \cellcolor{rowblue}16.7 & \cellcolor{rowblue}11.7 & \cellcolor{rowblue}10.5 \\
                & SFT-Symb-Direct                    & 19.7 & 0.3  & 11.9 \\
\bottomrule
\end{tabular}}
\caption{Stepwise soundness of various models under settings without inference-time interventions. The best variant of Llama and Qwen is highlighted.}
\label{tab:stepwise}
\vspace{-1em}
\end{table}

Table~\ref{tab:stepwise} reports the results of \textbf{stepwise soundness evaluation} across different models and training settings, offering a more fine-grained view of how well LLMs internalize logical reasoning principles.

The \textbf{All Valid} metric measures the proportion of samples where \emph{every} step is logically valid, a stringent indicator of formal reasoning. We observe that models fine-tuned with \textbf{SFT-NL} and \textbf{SFT-Symb-Struct} achieve high All Valid scores. Llama with SFT-NL reaches 40.9\%, substantially outperforming strong baselines like GPT-4o and DeepSeek-R1. The low score of DeepSeek-R1, despite its high accuracy, highlights a key insight from FineLogic. Having been trained solely on final-answer correctness, it often compresses multiple reasoning steps and omits necessary premises for intermediate conclusions. Since guessing the final label is easier than ensuring every inference is sound, accuracy alone can be misleading. "All Valid" enforces this stricter requirement, revealing a critical gap in the model's reasoning fidelity.

The \textbf{All Relevant} metric measures the proportion of samples in which every generated step is \emph{relevant}—i.e., none of the steps are redundant or unnecessary for reaching the conclusion. GPT-4o and LogiPT perform exceptionally well on this metric, implying that they rarely generate superfluous reasoning steps. In contrast, SFT-NL and SFT-Symb-Direct consistently underperform. For SFT-NL, this may stem from the nature of natural language reasoning: due to its semantic richness and lack of structural constraints, the model may occasionally include exploratory or overly verbose steps, unsure of which inference is most effective. For SFT-Symb-Direct, the poor performance is likely due to the model may failure to fully capture inter-fact dependencies, resulting in reasoning sequences that are logically valid but contain unused or irrelevant steps.

The \textbf{All Atomic} metric evaluates whether every step in a reasoning chain corresponds to a single atomic inference—i.e., whether steps avoid combining multiple logical moves. Here, \textbf{SFT-Symb-Struct} consistently outperforms other settings, highlighting the advantages of structured symbolic reasoning. Symbolic reasoning is inherently more compact and constrained, which likely helps the model learn what constitutes a minimal, rule-aligned inference step. In contrast, natural language reasoning often fuses multiple reasoning rules into a single step, making it harder for the model to isolate atomic operations.


\begin{table}[t]
\centering
\resizebox{\linewidth}{!}{%
\renewcommand{\arraystretch}{1.15}
\setlength{\tabcolsep}{6pt}
\definecolor{rowgreen}{HTML}{E2F0D9}  
\begin{tabular}{l|l c c c}
\toprule
\textbf{Model} & \textbf{Setting} & \textbf{CSS} & \textbf{RFI} & \textbf{NSD}\\
\midrule

\multirow{6}{*}{\centering \shortstack{Llama-3.1-\\8B-Instruct}}
                & \cellcolor{rowgreen}{\,--\,}        & \cellcolor{rowgreen}8.0 & \cellcolor{rowgreen}9.9 & \cellcolor{rowgreen}32.0 \\
                & LogiPT                              & 8.1 & 0.7 & 44.2 \\
                & \cellcolor{rowgreen}SFT-NL          & \cellcolor{rowgreen}8.5 & \cellcolor{rowgreen}9.9 & \cellcolor{rowgreen}\textbf{51.5} \\
                & SFT-Symb-Struct                     & 8.7 & 11.1 & 36.1 \\
                & \cellcolor{rowgreen}SFT-Symb-Filter & \cellcolor{rowgreen}\textbf{9.7} & \cellcolor{rowgreen}11.1 & \cellcolor{rowgreen}46.4 \\
                & SFT-Symb-Direct                     & 9.0 & \textbf{18.5} & 41.2 \\
\midrule
\multirow{6}{*}{\centering \shortstack{Qwen-2.5-\\7B-Instruct}}
                & \cellcolor{rowgreen}{\,--\,}        & \cellcolor{rowgreen}\textbf{8.6} & \cellcolor{rowgreen}7.4 & \cellcolor{rowgreen}43.3 \\
                & LogiPT                              & 8.1 & 9.2  & 43.2 \\
                & \cellcolor{rowgreen}SFT-NL          & \cellcolor{rowgreen}8.2 & \cellcolor{rowgreen}16.0 & \cellcolor{rowgreen}44.3 \\
                & SFT-Symb-Struct                     & 8.5 & 14.8 & 43.3 \\
                & \cellcolor{rowgreen}SFT-Symb-Filter & \cellcolor{rowgreen}8.3 & \cellcolor{rowgreen}16.0 & \cellcolor{rowgreen}\textbf{45.4} \\
                & SFT-Symb-Direct                     & 8.6 & \textbf{18.5} & 43.3 \\
\bottomrule
\end{tabular}}
\caption{Evaluation of Correctness Spanning Steps (CSS), Redundant Fact Identification (RFI), and Next-step Derivability (NSD) on Llama and Qwen. `-' indicates the original model. The best variant is highlighted.}
\label{tab:probing}
\vspace{-1em}
\end{table}





\subsubsection{Results on Representation-level Probing}
Table~\ref{tab:probing} presents results from our probing analysis, which assesses whether models internally acquire key reasoning abilities. Our findings reveal several key takeaways:

\textbf{Correctness Spanning Steps (CSS) shows a clear ceiling for 7/8B models under SFT.} Across all SFT variants and base models, the CSS score fluctuates by less than 1\%. This flat trend suggests that supervised fine-tuning alone offers little headroom for improving how early a model internally converges on the correct answer at this scale. Enhancing this "foresight" capability may require alternative methods like reinforcement learning with step-level rewards or architectural changes.

\textbf{Targeted supervision can unlock specific internal skills, as shown by Symb-Direct on RFI.} The \textbf{Redundant Fact Identification (RFI)} metric benefits most from the \textbf{SFT-Symb-Direct} setting. This training data uses minimal symbolization but retains redundant facts, forcing the model to learn to distinguish necessary from unnecessary premises. This direct alignment between the training task and the probing metric demonstrates that when a specific skill like redundancy filtering is desired, targeted supervision that mirrors the probe is highly effective.

\textbf{Next-Step Derivability (NSD) remains a challenge.} Performance on the \textbf{Next-Step Derivability (NSD)} task shows no consistent improvement across SFT variants. This indicates that vanilla SFT does not effectively enhance the model's internal representation of what logical steps are derivable next. This highlights a remaining gap in teaching explicit, forward-looking inference, suggesting promising future directions such as combining the NSD probe with self-supervised next-step prediction or curriculum-based RL.





\section{Future Work: Guiding Faithful Reasoning with Reinforcement Learning}

While supervised fine-tuning improves reasoning, reinforcement learning (RL) offers a path to refining it further. However, a major challenge in applying RL to logical reasoning is the reliance on sparse, binary rewards (i.e., whether the final answer is correct). This can lead to "reward hacking," where models learn heuristics to guess the right answer without developing a sound reasoning process.

The FineLogic framework offers a direct solution by providing dense, multi-faceted reward signals that can guide the quality of the reasoning process itself. Because our metrics are reference-free and deterministically scored, they translate cleanly into a multi-objective reward function.

\paragraph{A Multi-Objective Reward Framework.} The FineLogic metrics can be directly integrated into a granular, multi-objective reward function for RL. The \textbf{stepwise soundness} metrics (\textbf{validity}, \textbf{relevance}, \textbf{atomicity}) can be used to directly reward logical integrity, promoting reasoning that is rigorous, concise, and transparent. Concurrently, the representation-level metric \textbf{CSS} can serve as an online reward for cognitive efficiency, encouraging the model to converge on the correct answer earlier in its internal process. These components can be combined in a weighted sum:
\begin{align}
 R_{\text{total}} =& R_{\text{acc}} + w_v R_{\text{valid}} + w_r R_{\text{relevant}} \nonumber\\ &+ w_a R_{\text{atomic}} + w_{c} R_{\text{css}}   
\end{align}

In this framework, practitioners can tune the weights ($w_v, w_r, \dots$) to prioritize desired reasoning qualities like rigor or conciseness, offering a practical roadmap for training more faithful and interpretable models.

\section{Conclusion}

We introduce FineLogic, a unified and fine-grained framework for evaluating the logical reasoning capabilities of large language models. By integrating overall benchmark accuracy, stepwise soundness, and representation-level probing, FineLogic enables more interpretable and rigorous assessment beyond final-answer correctness. Leveraging this framework, we conduct a systematic investigation of how different fine-tuning supervision formats impact reasoning ability. Our experiments demonstrate that while natural language supervision leads to strong generalization and benchmark gains, symbolic styles better support minimal, rule-aligned reasoning structures. Furthermore, representation-level probing reveals that SFT primarily affects how models generate stepwise solutions rather than their ability to predict answers directly. These findings offer practical guidance for designing supervision strategies tailored to different reasoning objectives and highlight the importance of evaluating both behavioral and internal reasoning quality when advancing LLM reasoning systems. 
\newpage
\section*{Limitations}

Our work has two primary limitations. First, our evaluation is conducted on a fixed suite of datasets. While these were chosen for their diversity, this necessarily scopes our findings to these specific benchmarks, and the observed model behaviors may not generalize to the entire domain of logical reasoning tasks. Second, our analysis of supervision styles treats each format in isolation. Consequently, the potential benefits of hybrid training strategies, which could combine the complementary strengths of natural language and symbolic formats, remain unexplored in this study.

\section*{Acknowledgment}
We are grateful to Xiuying Chen of MBZUAI for her constructive comments and helpful suggestions, which helped improve our manuscript.

\bibliography{custom}

\newpage
\appendix

\section{Detailed Experimental Setup}
\label{app:detailed_setups}
\subsection{Detailed Dataset Information}
\label{app:datasets}

\paragraph{FLD} \cite{morishita2024enhancing}
This is a synthetic dataset designed to test generalization across varying reasoning depths. Our test set consists of 1,100 samples spanning 20 reasoning depths (0–19). It is specifically constructed to include out-of-distribution problems (depths 16–19) that are more difficult than those used for fine-tuning.

\paragraph{FOLIO} \cite{han2022folio}
A human-curated benchmark for first-order logic reasoning in natural language. We use the complete test set of 203 problems for our evaluation.

\paragraph{Multi-LogiEval} \cite{patel2024multi}
This dataset systematically combines multiple inference rules to create complex problems. To focus on challenging multi-step reasoning, we select 390 problems from its two most difficult subsets (depths 4 and 5), using only the First-Order and Propositional Logic categories.

\paragraph{ProntoQA} \cite{saparov2022language}
This dataset evaluates multi-hop reasoning over a semi-synthetic knowledge base. For our test set, we follow the challenging setup from prior work \cite{pan2023logic} and use the 500 hardest 5-hop problems. (The 3-hop problems used for fine-tuning are described in Section~\ref{sec:sft_data_design}).

\subsection{Detailed Baseline Methods}
\label{app:baseline}

\paragraph{LOGIPT} \cite{feng2023language} An approach where the language model is trained to directly generate symbolic reasoning steps, emulating a logical solver and bypassing an explicit natural language-to-symbol parsing stage.

\paragraph{Selection-Inference} \cite{creswell2022selection} A method that performs multi-step reasoning through an iterative process, alternating between a "selection" step to identify relevant facts from the context and an "inference" step to derive new conclusions.

\paragraph{LogicLM} \cite{pan2023logic} A neuro-symbolic framework where an LLM first translates a problem into a symbolic representation. A deterministic solver then performs the logical inference, and the LLM interprets the result back into natural language. It includes a self-refinement mechanism that uses solver feedback to correct translation errors.

\paragraph{SymbCoT} \cite{xu2024faithful} A framework that integrates symbolic logic into the Chain-of-Thought process. It translates the problem into a symbolic format and then generates reasoning steps using formal inference rules, often employing a verifier to check the logical soundness of the chain.




\section{Representation-Level Probing Implementation Details}
\label{app:probing}

We design three probing tasks to assess whether the model's internal representations capture reasoning-relevant information during multi-step logical problem solving. All probing experiments are conducted on a subset of the \textbf{FLD} dataset, specifically the 550 most complex problems requiring 10–20 reasoning steps. We use 450 problems for training and 100 for evaluation.

\subsection{Representation Extraction}

For all probing tasks, we extract the hidden state of the \textbf{final token from the last transformer layer} after processing the input prefix. The prefix consists of all reasoning steps up to a target step $k$ (i.e., steps 1 to $k$), and the final-token representation is treated as a summary of the model’s internal reasoning state at that point.

\subsection{Probing Model}

We use a lightweight yet effective classifier to probe the information contained in these hidden states. Specifically, we adopt a \textbf{logistic regression classifier} with feature standardization and \textbf{5-fold cross-validation} for hyperparameter selection. This setup ensures a simple and interpretable linear decision boundary while maintaining robustness against overfitting. The classifier is trained solely on the extracted representations, while the underlying language model remains frozen throughout the probing process.

\subsection{Task 1: Correctness Spanning Steps}

This task evaluates how early in the reasoning process the model internalizes the correct final answer. For a problem requiring $n$ reasoning steps, we:

\begin{itemize}
    \item Generate $n$ input prefixes, each ending at step $i$, where $i \in [1, n]$.
    \item Train a probing classifier to predict the ground-truth label (True / False) based on the representation at each prefix.
    \item For each test sample, we identify the smallest $i$ such that the classifier correctly predicts the label at step $i$ but fails at step $i-1$.
\end{itemize}

The \textbf{correctness spanning length} is defined as $n - i$, capturing how early the model “knows” the correct answer.

\subsection{Task 2: Redundant Facts Identification}

This task assesses whether the model can distinguish between relevant and irrelevant facts. For each sample:

\begin{itemize}
    \item We locate the point after all facts and the hypothesis have been presented.
    \item We construct six variants of the input: three with \textbf{necessary facts} (used later in the proof), and three with \textbf{redundant facts} (unused in any proof step).
    \item The classifier is trained to predict whether the appended facts are necessary or redundant based on the updated representation.
\end{itemize}

This task tests whether the model encodes awareness of which premises are logically relevant for solving the task.

\subsection{Task 3: Next-Step Derivability}

This task probes whether the model can determine which steps are logically available at a given point in the proof. For each sample:

\begin{itemize}
    \item We randomly select six intermediate steps.
    \item At each step, we append three \textbf{valid next steps} (that are inferable from the current context) and three \textbf{invalid steps} (that appear later in the proof but are not yet derivable).
    \item The classifier is trained to distinguish between currently valid and invalid steps.
\end{itemize}

This task evaluates whether the model has encoded an implicit understanding of the forward progression of logical inference.

\section{Additional Experiments}
\label{app:experiment_details}

\begin{table*}[t]
\centering
\renewcommand{\arraystretch}{1}
\scalebox{0.75}{%
\begin{tabular}{cc P{2.2cm} P{2.2cm} P{2.2cm} P{2.2cm}}
\toprule
\multirow{2}{*}{\textbf{Model}} & \multirow{2}{*}{\textbf{Setting}} & \multirow{2}{*}{\textbf{FLD}} & \multirow{2}{*}{\textbf{FOLIO}} & \multirow{2}{*}{\makecell[c]{\textbf{Multi-}\\\textbf{LogiEval}}} & \multirow{2}{*}{\makecell[c]{\textbf{ProntoQA}}} \\
& & & & & \\
\midrule
\rowcolor{gray!10}
& Few-shot         & 7.4  & 30.5 & 0.5  & 0.2 \\
\rowcolor{gray!10}
& SFT-NL           & \textbf{58.8} & \textbf{56.2} & 45.6 & \textbf{97.0} \\
\rowcolor{gray!10}
& SFT-Symb-Struct  & 38.9 & 49.3 & \textbf{60.3} & 90.6 \\
\rowcolor{gray!10}
& SFT-Symb-Filter  & 56.1 & 50.2 & 59.2 & 91.4 \\
\rowcolor{gray!10} \multirow{-5}{*}{\textbf{\shortstack{Qwen-2.5-7B-\\Math-Instruct}}}
& SFT-Symb-Direct  & 47.9 & 43.8 & 39.5 & 51.4 \\
\addlinespace
& Few-shot         & 44.0 & \textbf{81.8} & \textbf{73.6} & \textbf{99.8} \\
& SFT-NL           & \textbf{75.4} & 64.0 & 48.7 & 99.4 \\
& SFT-Symb-Struct  & 58.6 & 59.1 & 61.0 & 99.6 \\
& SFT-Symb-Filter  & 63.6 & 57.1 & 64.1 & \textbf{99.8} \\
\multirow{-5}{*}{\textbf{\shortstack{Qwen-3-\\8B}}}
& SFT-Symb-Direct  & 54.6 & 56.2 & 63.1 & 75.2 \\
\bottomrule
\end{tabular}
}
\caption{Overall benchmark accuracy on \textbf{Qwen-2.5-7B-Math-Instruct} and \textbf{Qwen-3-8B} across training settings. Per-model column-wise maxima are in \textbf{bold}.}
\label{apptab:overall}
\end{table*}

\begin{table}[t]
\centering
\resizebox{\linewidth}{!}{%
\renewcommand{\arraystretch}{1.15}
\setlength{\tabcolsep}{6pt}
\begin{tabular}{l|lrrr}
\toprule
\textbf{Model} & \textbf{Setting} & \textbf{All Valid} & \textbf{All Relevant} & \textbf{All Atomic}\\
\midrule
\multirow{5}{*}{\centering \shortstack{Qwen-2.5-7B-\\Math-Instruct}}
                & \cellcolor{rowblue}Few-shot        & \cellcolor{rowblue}4.5  & \cellcolor{rowblue}\textbf{65.8} & \cellcolor{rowblue}3.9 \\
                & SFT-NL                              & 6.5  & 60.4 & 2.0 \\
                & \cellcolor{rowblue}SFT-Symb-Struct & \cellcolor{rowblue}20.3 & \cellcolor{rowblue}21.7 & \cellcolor{rowblue}11.0 \\
                & SFT-Symb-Filter                     & 12.4 & 3.6  & 5.6 \\
                & \cellcolor{rowblue}SFT-Symb-Direct           & \cellcolor{rowblue}\textbf{25.9} & \cellcolor{rowblue}0.5  & \cellcolor{rowblue}\textbf{14.8} \\
\midrule
\multirow{5}{*}{\centering \shortstack{Qwen-3-\\8B}}
                & \cellcolor{rowblue}Few-shot        & \cellcolor{rowblue}12.1 & \cellcolor{rowblue}49.5 & \cellcolor{rowblue}6.5 \\
                & SFT-NL                    & 17.3 & \textbf{84.5} & 3.5 \\
                & \cellcolor{rowblue}SFT-Symb-Struct & \cellcolor{rowblue}\textbf{26.1} & \cellcolor{rowblue}5.5  & \cellcolor{rowblue}\textbf{13.0} \\
                & SFT-Symb-Filter                     & 19.1 & 2.3  & 7.6 \\
                & \cellcolor{rowblue}SFT-Symb-Direct                     & \cellcolor{rowblue}24.7 & \cellcolor{rowblue}0.4  & \cellcolor{rowblue}12.8 \\
\bottomrule
\end{tabular}}
\caption{Stepwise soundness after \texttt{</think>} on \textbf{Qwen-2.5-7B-Math-Instruct} and \textbf{Qwen-3-8B}. Per-model column-wise maxima are in \textbf{bold}.}
\label{apptab:stepwise}
\vspace{-1em}
\end{table}

\definecolor{rowgreen}{HTML}{E2F0D9}
\begin{table}[t]
\centering
\resizebox{\linewidth}{!}{%
\renewcommand{\arraystretch}{1.15}
\setlength{\tabcolsep}{6pt}
\begin{tabular}{l|l c c c}
\toprule
\textbf{Model} & \textbf{Setting} & \textbf{CSS} & \textbf{RFI} & \textbf{NSD}\\
\midrule
\multirow{5}{*}{\centering \shortstack{Qwen-2.5-7B-\\Math-Instruct}}
                & \cellcolor{rowgreen}{\,--\,}        & \cellcolor{rowgreen}7.4 & \cellcolor{rowgreen}4.9  & \cellcolor{rowgreen}37.1 \\
                & SFT-NL                     & 8.6 & 6.2  & \textbf{42.3} \\
                & \cellcolor{rowgreen}SFT-Symb-Struct                     & \cellcolor{rowgreen}7.8 & \cellcolor{rowgreen}9.9  & \cellcolor{rowgreen}36.1 \\
                & SFT-Symb-Filter                     & 8.3 & 8.6  & 37.1 \\
                & \cellcolor{rowgreen}SFT-Symb-Direct            & \cellcolor{rowgreen}\textbf{8.7} & \cellcolor{rowgreen}\textbf{11.6} & \cellcolor{rowgreen}35.1 \\
\midrule
\multirow{5}{*}{\centering \shortstack{Qwen-3-\\8B}}
                & \cellcolor{rowgreen}\textbf{--}     & \cellcolor{rowgreen}\textbf{10.0} & \cellcolor{rowgreen}11.1 & \cellcolor{rowgreen}30.9 \\
                & SFT-NL                     & 9.8 & \textbf{23.5} & 36.1 \\
                & \cellcolor{rowgreen}SFT-Symb-Struct                     & \cellcolor{rowgreen}9.9 & \cellcolor{rowgreen}9.9  & \cellcolor{rowgreen}32.0 \\
                & SFT-Symb-Filter & \textbf{10.0} & 14.8 & \textbf{39.2} \\
                & \cellcolor{rowgreen}SFT-Symb-Direct                     & \cellcolor{rowgreen}9.7 & \cellcolor{rowgreen}17.3 & \cellcolor{rowgreen}35.1 \\
\bottomrule
\end{tabular}}
\caption{Representation-level probing (CSS, RFI, NSD) on \textbf{Qwen-2.5-7B-Math-Instruct} and \textbf{Qwen-3-8B}. Per-model column-wise maxima are in \textbf{bold}; the \texttt{--} row denotes the original model.}
\label{apptab:probing}
\vspace{-1em}
\end{table}

This section provides further details and aggregate analysis of our experimental results.

\paragraph{Aggregate Findings Across Models.} To broaden our findings, we evaluated two additional models, Qwen-2.5-Math-7B and Qwen-3-8B. The results, shown in Tables \ref{apptab:overall}, \ref{apptab:stepwise} and \ref{apptab:probing}, combined with our primary experiments, yield several key aggregate takeaways:

\begin{itemize}[leftmargin=*,itemsep=2pt,parsep=0pt]
    \item \textbf{Natural-language SFT is the most reliable path to higher accuracy.} Across all four base models and four datasets, SFT-NL consistently delivers the largest and most robust gains in final-answer correctness, reaffirming that aligning with the model's pre-training modality is most effective for generalization.

    \item \textbf{Symbolic curricula with redundant facts improve chain integrity.} SFT settings that retain logically valid but extraneous premises (Symb-Direct and Symb-Struct) tend to improve \textit{All-Valid} and \textit{All-Atomic} scores. This suggests that exposing models to "noisy but sound" proof environments encourages more careful and explicit rule application.

    \item \textbf{CSS is more sensitive to architecture and RL than to SFT.} The larger, RL-enhanced Qwen-3-8B model achieves a notably higher base CSS score than other models. However, its CSS remains largely unchanged by SFT, providing strong evidence that deeper latent reasoning depends more on model design and training regimes like RL, rather than supervised instruction tuning.
\end{itemize}

\paragraph{Detailed Analysis of Performance by Reasoning Depth.}
A more granular breakdown of performance by reasoning depth, presented in \autoref{tab:step_results}, reveals further nuances in these trends. While models fine-tuned with natural language supervision (e.g., Llama-3.1-SFT-NL achieving 89.5\% accuracy for 0-3 steps on FLD) perform strongly on tasks with shallower reasoning depths, their symbolic reasoning counterparts tend to exhibit greater resilience as the complexity and number of reasoning steps increase. For instance, on FLD problems requiring 16-19 steps, Llama-3.1-SFT-Symb-Filter (62.5\%) and Llama-3.1-SFT-Symb-Struct (58.5\%) maintain higher accuracy compared to Llama-3.1-SFT-NL (46.0\%), highlighting the benefit of symbolic formats for robust long-chain inference.

\begin{table*}[htbp]
\centering
\scalebox{0.85}{%
\begin{tabular}{cc P{1.6cm} P{1.6cm} P{1.6cm} P{1.6cm} P{1.6cm}}
\toprule
\multirow{2}{*}{\textbf{Model}} & \multirow{2}{*}{\textbf{Setting}} & \multicolumn{5}{c}{\textbf{FLD Accuracy by Step}} \\
\cmidrule(lr){3-7}
& & 0--3 & 4--7 & 8--11 & 12--15 & 16--19 \\
\midrule
\rowcolor{gray!10}
& Direct     & 78.5 & 63.0 & 43.0 & 46.5 & 35.5 \\
\rowcolor{gray!10}
& CoT        & 82.0 & 62.0 & 56.5 & 44.0 & 46.5 \\
\rowcolor{gray!10}
& Few-shot   & 81.5 & 68.5 & 53.5 & 46.0 & 47.0 \\
\rowcolor{gray!10}
& Logic-LM   & 68.4 & 52.1 & 31.5 & 28.2 & 22.8 \\
\rowcolor{gray!10}
& SymbCoT    & 69.9 & 52.5 & 32.0 & 26.5 & 24.5 \\
\rowcolor{gray!10} \multirow{-6}{*}{\textbf{GPT-4o}}
& Sel-Inf    & 64.5 & 55.5 & 49.5 & 49.0 & 55.5 \\
\addlinespace
& Direct     & 92.5 & 86.0 & 80.5 & 75.0 & 76.5 \\
& CoT        & 92.0 & 86.0 & 78.0 & 77.5 & 73.5 \\
& Few-shot   & 89.0 & 85.0 & 80.5 & 69.0 & 71.0 \\
& Logic-LM   & 91.4 & 78.6 & 64.8 & 58.2 & 52.4 \\
& SymbCoT    & 86.4 & 80.5 & 70.9 & 45.2 & 53.4 \\ \multirow{-6}{*}{\textbf{\shortstack{DeepSeek\\-R1}}}
& Sel-Inf    & 93.0 & 88.0 & 84.5 & 79.0 & 75.5 \\
\addlinespace
\rowcolor{gray!10}
& Direct     & 40.5 & 30.0 & 24.0 & 27.0 & 25.5 \\
\rowcolor{gray!10}
& CoT        & 41.5 & 32.0 & 29.0 & 24.0 & 19.5 \\
\rowcolor{gray!10}
& Few-shot   & 49.5 & 45.5 & 33.0 & 39.0 & 32.0 \\
\rowcolor{gray!10}
& Logic-LM   & 56.4 & 41.3 & 32.6 & 28.8 & 26.2 \\
\rowcolor{gray!10}
& SymbCoT    & 57.8 & 41.0 & 39.0 & 37.8 & 35.4 \\
\rowcolor{gray!10}
& Sel-Inf    & 63.5 & 55.5 & 52.5 & 45.0 & 42.0 \\
\rowcolor{gray!10}
& LogiPT     & 72.5 & 53.5 & 51.0 & 35.0 & 37.0 \\
\rowcolor{gray!10}
& SFT-NL     & 89.5 & 72.5 & 52.0 & 56.5 & 46.0 \\
\rowcolor{gray!10}
& SFT-Symb-Struct & 88.5 & 78.5 & 65.0 & 66.0 & 58.5 \\
\rowcolor{gray!10}
& SFT-Symb-Filter & 72.0 & 73.0 & 67.5 & 72.0 & 62.5 \\
\rowcolor{gray!10} \multirow{-11}{*}{\textbf{\shortstack{Llama-3.1\\-8B-Instruct}}}
& SFT-Symb-Direct & 81.0 & 58.5 & 48.5 & 38.5 & 27.5 \\
\addlinespace
& Direct     & 69.0 & 45.5 & 45.0 & 38.5 & 36.0 \\
& CoT        & 70.5 & 55.5 & 36.5 & 42.5 & 40.5 \\
& Few-shot   & 63.0 & 44.0 & 33.5 & 27.0 & 33.0 \\
& Logic-LM   & 68.7 & 51.2 & 31.4 & 26.0 & 29.2 \\
& SymbCoT    & 52.3 & 39.5 & 30.7 & 28.1 & 19.9 \\
& Sel-Inf    & 49.0 & 26.5 & 29.5 & 27.0 & 24.5 \\
& LogiPT     & 80.5 & 74.0 & 64.0 & 68.0 & 57.5 \\
& SFT-NL     & 93.5 & 78.5 & 67.5 & 66.5 & 52.5 \\
& SFT-Symb-Struct & 74.5 & 65.5 & 50.0 & 43.0 & 34.0 \\
& SFT-Symb-Filter & 68.0 & 59.5 & 48.0 & 51.5 & 45.0 \\ \multirow{-11}{*}{\textbf{\shortstack{Qwen-2.5\\-7B-Instruct}}}
& SFT-Symb-Direct & 82.0 & 60.0 & 51.5 & 42.5 & 37.0 \\
\bottomrule
\end{tabular}
}
\caption{FLD accuracy breakdown by reasoning step ranges}
\label{tab:step_results}
\end{table*}

\section{Computational Resources}
All supervised fine-tuning experiments were conducted using 4 NVIDIA A100 GPUs. Each model was trained for approximately 2 hours. Evaluation on the full suite of benchmarks and diagnostic metrics required an additional 0.5 hours per model.

\section{Example and Case Study}
\label{app:training_examples}
\begin{figure*}[t]
    \centering
    \includegraphics[width=\linewidth]{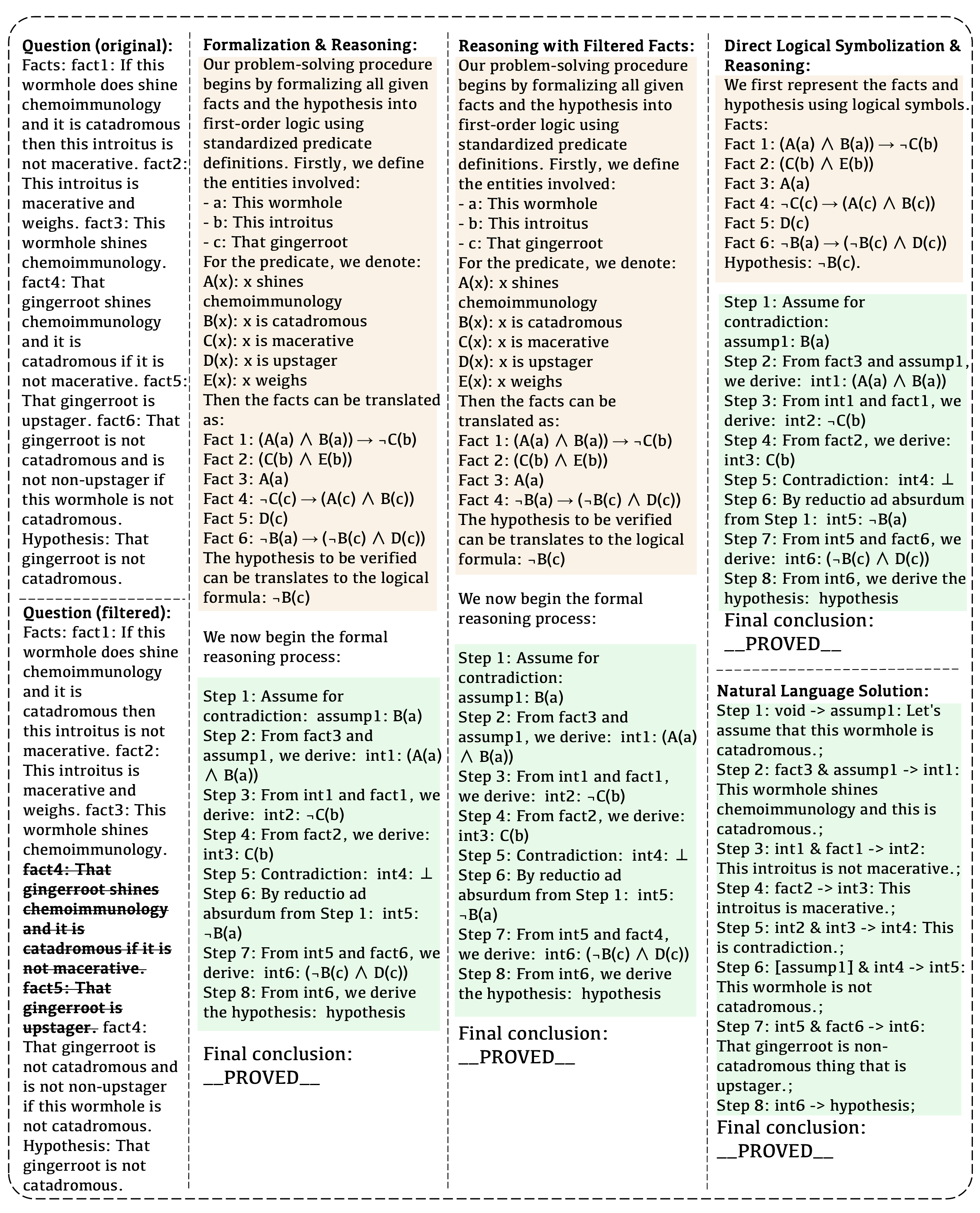} 
    \caption{Comparison of a logical reasoning problem under four distinct training data settings. The figure illustrates: (a) direct logical symbolization and reasoning ; (b) full formalization in first-order logic, including definitions and fact translation ; (c) reasoning conducted purely in natural language; and (d) formal reasoning using a pre-filtered set of facts. This comparison highlights the differences in processing pathways and the structure of the resulting solutions for each approach.}
    \label{fig:different_training_case}
\end{figure*}

\begin{figure*}[t]
    \centering
    \includegraphics[width=\linewidth]{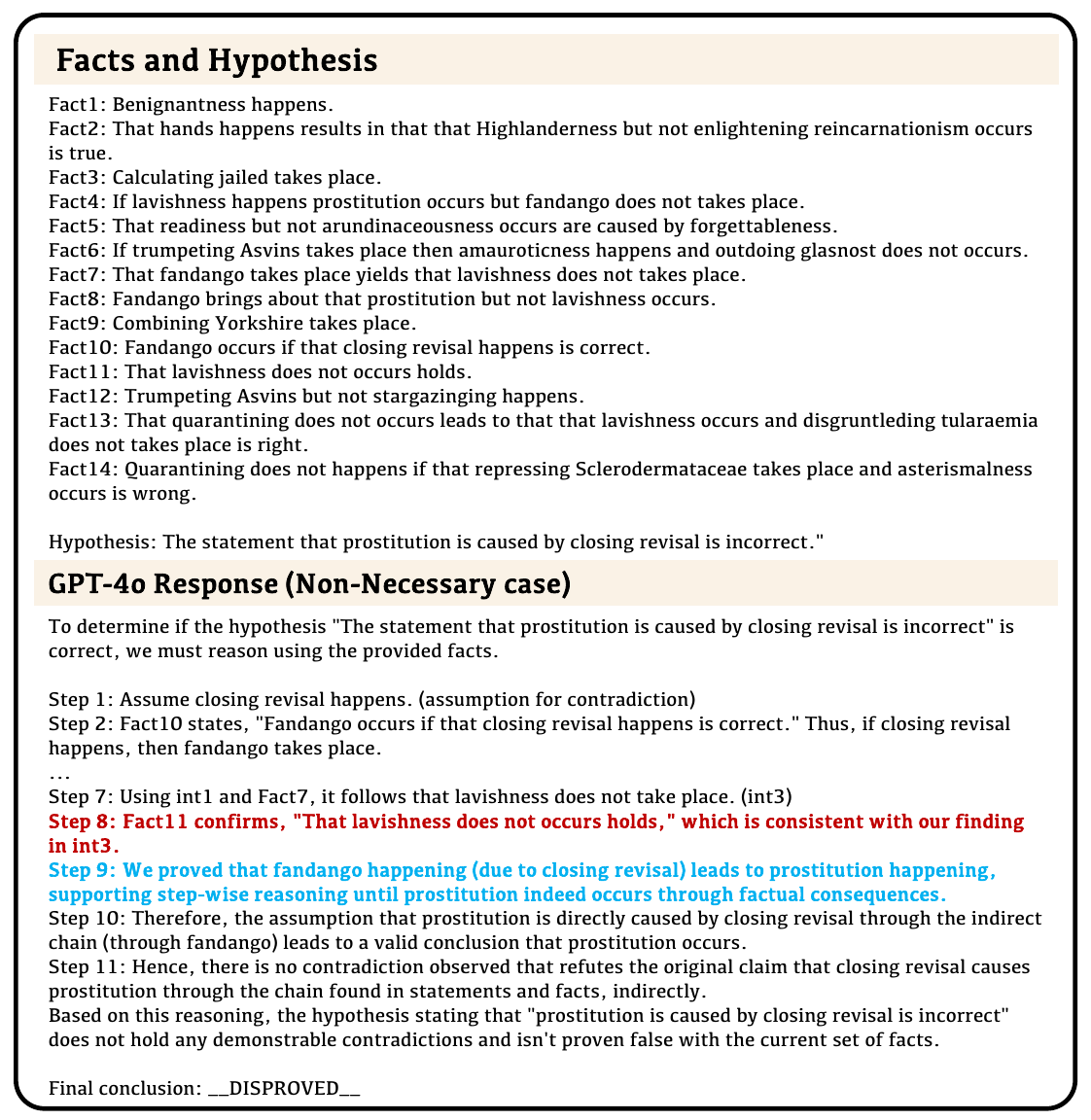} 
    \caption{Case Study: In the GPT-4o response, Step 8 redundantly confirms the conclusion from Step 7 (int3) by citing Fact11, making it an unnecessary step in the reasoning chain. Step 9 is not atomic because it summarizes a multi-step inference (from "fandango happening" to "prostitution happening" using Fact 8) as a single proven outcome without detailing the individual logical operations involved.}
    \label{fig:casestudy_ness}
\end{figure*}

This section showcases examples from our training dataset along with an error case study. Further details can be found in \autoref{fig:different_training_case} and \autoref{fig:casestudy_ness}.

\section{Prompt Template}
\label{app:prompt_template}
This section showcases various prompts, encompassing those designed for reasoning and data generation, as detailed in Figures \ref{prompt:DirectReasoning},\ref{prompt:CoTReasoning},\ref{prompt:FewShotReasoning},\ref{prompt:ExtractEntitiesPredicates},\ref{prompt:ExtractPredicatesOnly},\ref{prompt:LogicProofTranslation},\ref{prompt:LogicalProofGeneration},\ref{prompt:StepValidity},\ref{prompt:StepAtomicity}

\section{Use of AI Assistants in Manuscript Preparation}
During the preparation of this manuscript, we utilized large language models, including GPT and Gemini, as writing assistants. Their role was limited to improving the clarity, grammar, and readability of the text. This included tasks such as rephrasing sentences, correcting spelling errors, and ensuring stylistic consistency. The core scientific contributions, experimental design, and analysis presented in this paper are solely the work of the human authors, who take full responsibility for the final content.

\begin{figure*}[h!]
\centering
\begin{bluebox}[
Prompt Template: Direct Reasoning
]
\texttt{
Based on the provided facts, answer the question. Conclude with one of the markers: "\_\_PROVED\_\_" for proven, "\_\_DISPROVED\_\_" for disproven, or "\_\_UNKNOWN\_\_" if uncertain.
\newline Facts:\{facts\}
\newline Hypothesis:\{hypothesis\}
}

\end{bluebox}
\caption{Prompt template for direct reasoning. Placeholders: \texttt{\{facts\}}, \texttt{\{hypothesis\}}.}
\label{prompt:DirectReasoning}
\end{figure*}

\begin{figure*}[h!]
\centering
\begin{bluebox}[
Prompt Template: CoT Reasoning
]
\texttt{
Based on the provided facts, answer the question. Conclude with one of the markers: "\_\_PROVED\_\_" for proven, "\_\_DISPROVED\_\_" for disproven, or "\_\_UNKNOWN\_\_" if uncertain.
\newline Facts:\{facts\}
\newline Hypothesis:\{hypothesis\}
\newline Let's analyze this step by step.
}
\end{bluebox}
\caption{Prompt template for Chain-of-Thought (CoT) reasoning. Placeholders: \texttt{\{facts\}}, \texttt{\{hypothesis\}}.}
\label{prompt:CoTReasoning}
\end{figure*}

\begin{figure*}[h!]
\centering
\begin{bluebox}[
Prompt Template: Few-Shot Reasoning
]
\texttt{
Based on the provided facts, answer the question. Conclude with one of the markers: "\_\_PROVED\_\_" for proven, "\_\_DISPROVED\_\_" for disproven, or "\_\_UNKNOWN\_\_" if uncertain.
\newline Here are some examples of proofs for your reference:
\newline [Start of example]
\newline For example, for this question:
\newline \{example\}
\newline [End of example]
\newline You can refer to the proof method of the above question, think step by step, and give the result of this question.
\newline Facts:\{facts\}
\newline Hypothesis:\{hypothesis\}
}
\end{bluebox}
\caption{Prompt template for few-shot reasoning. Placeholder: \texttt{\{example\}}, \texttt{\{facts\}}, \texttt{\{hypothesis\}}..}
\label{prompt:FewShotReasoning}
\end{figure*}

\begin{figure*}[h!]
\centering
\begin{bluebox}[
Prompt Template: Entity and Predicate Extraction
]
\texttt{
You are a logic analysis expert. Please extract all entities and predicates from the following logical expression translations:
\newline Translation content: \{formula\_translations\}
\newline facts\_formula: \{facts\_formula\}
\newline facts: \{facts\}
\newline Special Requirement: If any entity or predicate symbol appears in the facts\_formula, but has NO direct definition in the Translation content, you MUST go to the facts section and locate the corresponding natural language description and extract it. Be extremely careful NOT to omit any such entities or predicates. Only skip if it is literally missing from both translation content and facts.
\newline Task:
\newline 1. Identify all entities involved (e.g., this tablefork, this corsair) and assign variables to them (a, b, c, d...)
\newline 2. Identify all predicates (e.g., is a raised, is a collotype) and assign symbols (using the original symbols like A, B, C...)
\newline Critical instructions:
\newline - Only give full entity and predicate explanations if their definitions appear in the formula\_translations or facts.
\newline - Only include entities and predicates that explicitly appear in the provided translation content or facts.
\newline - Do not invent, infer, or add any entities or predicates not directly mentioned in the translations or facts.
\newline - Maintain the original variable identifiers (e.g., 'a' in A(a) corresponds to the first entity).
\newline - Maintain the original predicate identifiers (e.g., 'A' in A(x) represents "x is a raised").
\newline - If a symbol (like 'c', 'F', etc.) doesn't appear in the translations or facts, do not include it in your output.
\newline Expected output format:
\newline We define the entities involved:
\newline - a: [Corresponding entity, e.g., "This tablefork"]
\newline - b: [Corresponding entity, e.g., "This corsair"]...
\newline We denote:
\newline [Original predicate symbol](x): [Predicate description]
\newline [Original predicate symbol](x): [Predicate description]...
\newline Please provide only the requested definitions without any additional information or explanations.
}
\end{bluebox}
\caption{Prompt template for extracting entities and predicates when lowercase variables (entities) are present. Placeholders: \texttt{\{formula\_translations\}}, \texttt{\{facts\_formula\}}, \texttt{\{facts\}}.}
\label{prompt:ExtractEntitiesPredicates}
\end{figure*}

\begin{figure*}[h!]
\centering
\begin{bluebox}[
Prompt Template: Predicate Extraction (No Entities)
]
\texttt{
You are a logic analysis expert. Please extract all predicates from the following logical expression translations:
\newline Translation content: \{formula\_translations\}
\newline facts\_formula: \{facts\_formula\}
\newline facts: \{facts\}
\newline Special Requirement: If any entity or predicate symbol appears in the facts\_formula, but has NO direct definition in the Translation content, you MUST go to the facts section and locate the corresponding natural language description and extract it. Be extremely careful NOT to omit any such entities or predicates. Only skip if it is literally missing from both translation content and facts.
\newline Task: Identify all predicates and translate each uppercase symbol directly.
\newline Critical instructions:
\newline - For each uppercase symbol in the facts\_formula, provide a direct translation in the format: [SYMBOL]: xxx happened.
\newline - **Do not omit any symbols that appear in facts\_formula or translation content. If they appear, they must be translated.**
\newline - Only include symbols that actually appear in the facts\_formula or translation content.
\newline - Do not invent or infer any entities or relationships not explicitly mentioned.
\newline - If a predicate's meaning is clearly defined in the translations or facts, use that definition.
\newline - Do not include any lowercase symbols or entity definitions as they are not relevant in this case.
\newline - If some symbols appear in facts\_formula but not in translation content, you can directly translate the entire formula expression containing those symbols rather than translating each symbol individually. For example, for an expression like \ensuremath{\neg}C\ensuremath{\rightarrow}\ensuremath{\neg}(F\ensuremath{\land}\ensuremath{\neg}E), you don't need to separately translate E if it's not defined elsewhere.
\newline Expected output format:
\newline We define:
\newline A: xxx happened.
\newline B: xxx happened.
\newline AB: xxx happened...
\newline Please provide only the requested definitions without any additional information or explanations.
}
\end{bluebox}
\caption{Prompt template for extracting predicates when no lowercase variables (entities) are present. Placeholders: \texttt{\{formula\_translations\}}, \texttt{\{facts\_formula\}}, \texttt{\{facts\}}.}
\label{prompt:ExtractPredicatesOnly}
\end{figure*}

\begin{figure*}[h]
\centering
\begin{bluebox}[
Prompt Template: Logic Proof Translation
]
\texttt{
You are a logic proof translator. Your task is to translate a logical proof sequence from symbolic notation into a clear, step-by-step explanation.
\newline Given: 1. A proof sequence in symbolic form 2. Definitions of entities and predicates used in the proof 3. Logical formula translations
\newline Task: Convert the symbolic proof into a concise, step-by-step explanation that a human can easily follow.
\newline Proof sequence to translate: \{proofs\_sentence\}
\newline Conclusion: \{conclusion\}
\newline Instructions for translation:
\newline 1. Split the proof at each semicolon (;) to identify individual steps.
\newline 2. For each step: First, write a brief, natural language explanation on its own line (e.g. "Assume for contradiction: [formula]" or "From [inputs], we derive:"). On the next line, write the step label and the logical formula as in the original proof (e.g. assump1: A(b), int2: ¬B(b), etc.). Do not put both the explanation and the formula on the same line. For assumptions, use "Assume for contradiction: [formula]" then write assumpX: [formula] on the following line. For a standard derived step, use "From [inputs], we derive:" then on the following line write intX: [formula]. For contradictions, use "Contradiction:" then on the following line write "⊥". For reductio ad absurdum, use "By reductio ad absurdum from [step number]:" then write the derived conclusion on the next line. Do not skip formula labels or step names. Write both the explanation and the labeled formula.
\newline 3. Maintain correct logical notation (such as ¬, ∧, ∨, →, ∃, ⊥, etc.).
\newline 4. In the final step, clearly relate the conclusion to the hypothesis, if appropriate.
\newline 5. The output should be only the formatted translation, with no additional commentary.
\newline Output format:
\newline Step 1: [Brief explanation]
\newline [Formula derived]
\newline Step 2: From [input], we derive:
\newline [Formula derived]
\newline Step 3: Assume for contradiction:
\newline assumpX: [Formula derived]
\newline ...
\newline \{status\_message\_content\}
\newline Final conclusion: \{conclusion\}
\newline The conclusion must use exactly two underscores before and after either PROVED or DISPROVED or UNKNOWN, with no additional spaces or characters. Translate the proof concisely but retain all logical information from the original proof sequence. Do not add any steps not present in the original, and do not skip any steps. Output the translation only, with no additional commentary.
}
\end{bluebox}
\caption{Prompt template for logic proof translation. The placeholder \texttt{\{proofs\_sentence\}} is for the symbolic proof sequence. The placeholder \texttt{\{conclusion\}} is for the conclusion (\_\_PROVED\_\_/\_\_DISPROVED\_\_/\_\_UNKNOWN\_\_). The placeholder \texttt{\{status\_message\_content\}} is replaced by the string 'The search path has been exhausted without finding a way to either prove or disprove the hypothesis.' if \texttt{\{conclusion\}} is '\_\_UNKNOWN\_\_', and is an empty string otherwise (which will result in different spacing around it as per the original prompt generation logic).}
\label{prompt:LogicProofTranslation}
\end{figure*}

\begin{figure*}[h!]
\centering
\begin{bluebox}[
Prompt Template: Logical Proof Generation
]
\texttt{Solve the following logical reasoning problem using formal symbolic logic and provide a step-by-step reasoning process.
\newline Follow these steps precisely:
\newline 1. Define predicates to represent terms in the problem
\newline 2. Translate all facts and the hypothesis into formal logical expressions
\newline 3. Derive the conclusion through systematic reasoning
\newline 4. State the final conclusion
\newline OUTPUT FORMAT:
\newline Your answer should follow this format exactly:
\newline - Begin with "Our problem-solving procedure begins by formalizing all given facts and the hypothesis into first-order logic using standardized predicate definitions."
\newline - Then state "For the predicate, we denote:" followed by your predicate definitions
\newline - Translate each fact into a formal logical expression
\newline - Present your reasoning steps in numbered format (Step 1:, Step 2:, etc.)
\newline - End with "Final conclusion: " followed by either "\_\_PROVED\_\_" or "\_\_DISPROVED\_\_"
\newline IMPORTANT: The conclusion must use exactly two underscores before and after either PROVED or DISPROVED, with no additional spaces or characters.
\newline Here is an example problem solution, You need to strictly follow the format like this:
\newline Example Solution:
\newline \{fewshot\_example\}
\newline Now, solve this problem: \{question\}
\newline The answer should be: \{label\}
\newline Provide only the solution with no additional commentary or preamble.
}
\end{bluebox}
\caption{Prompt template for generating a logical reasoning process. Placeholders: \texttt{\{question\}} for the problem statement, \texttt{\{label\}} for the expected answer (e.g., "\_\_PROVED\_\_"), and \texttt{\{fewshot\_example\}} for a formatted example solution.}
\label{prompt:LogicalProofGeneration}
\end{figure*}

\begin{figure*}[h!]
\centering
\begin{bluebox}[
Prompt Template: Step Validity Evaluation
]
\texttt{
Premises:\newline
\{premises\_str\}\newline
\newline
Conclusion:\newline
\{concl\_text\_full\}\newline
\newline
Do the premises entail the conclusion? Answer true or false only.
}
\end{bluebox}
\caption{Prompt template for evaluating step validity. Placeholders: \texttt{\{premises\_str\}} (a string listing the premises, e.g., "fact1: Text of fact 1 int1: Text of intermediate 1"), \texttt{\{concl\_text\_full\}} (a string representing the conclusion, e.g., "int2: Text of intermediate 2" or "hypothesis: Text of hypothesis"). The model is expected to return 'true' or 'false'.}
\label{prompt:StepValidity}
\end{figure*}

\begin{figure*}[h!]
\centering
\begin{bluebox}[
Prompt Template: Step Atomicity Evaluation
]
\texttt{
Premises:\newline
\{premises\_str\}\newline
\newline
Conclusion:\newline
\{concl\_text\_full\}\newline
\newline
Is this inference atomic...? Answer true or false only.
}
\end{bluebox}
\caption{Prompt template for evaluating step atomicity. Placeholders: \texttt{\{premises\_str\}} (a string listing the premises), \texttt{\{concl\_text\_full\}} (a string representing the conclusion). The model is expected to return 'true' or 'false' indicating if the inference from premises to conclusion is a single, indivisible logical step.}
\label{prompt:StepAtomicity}
\end{figure*}

\end{document}